\documentclass{bmvc2k}

\usepackage{times}
\usepackage{epsfig}
\usepackage{graphicx}
\usepackage{amsmath}
\usepackage{amssymb}

\usepackage{makecell}
\usepackage{mathtools}
\usepackage{capt-of}

\title{CornerNet-Lite: Efficient Keypoint-Based Object Detection}

\addauthor{Hei Law}{heilaw@cs.princeton.edu}{1}
\addauthor{Yun Tang}{yteng@princeton.edu}{1}
\addauthor{Olga Russakovsky}{olgarus@cs.princeton.edu}{1}
\addauthor{Jia Deng}{jiadeng@cs.princeton.edu}{1}

\addinstitution{
    Department of Computer Science \\
    Princeton University \\
    Princeton, NJ USA
}

\runninghead{H. Law, Y. Tang, O. Russakovsky, J. Deng}{CornerNet-Lite}

\begin{document}

\maketitle

\begin{abstract}
Keypoint-based methods are a relatively new paradigm in object detection, eliminating the need for anchor boxes and offering a simplified detection framework. Keypoint-based CornerNet achieves state of the art accuracy among single-stage detectors. However, this accuracy comes at high processing cost. In this work, we tackle the problem of efficient keypoint-based object detection and introduce CornerNet-Lite. CornerNet-Lite is a combination of two efficient variants of CornerNet: CornerNet-Saccade, which uses an attention mechanism to eliminate the need for exhaustively processing all pixels of the image, and CornerNet-Squeeze, which introduces a new compact backbone architecture. Together these two variants address the two critical use cases in efficient object detection: improving efficiency without sacrificing accuracy, and improving accuracy at real-time efficiency. CornerNet-Saccade is suitable for offline processing, improving the efficiency of CornerNet by 6.0x and the AP by 1.0\% on COCO. CornerNet-Squeeze is suitable for real-time detection, improving both the efficiency and accuracy of the popular real-time detector YOLOv3 (34.4\% AP at 30ms for CornerNet-Squeeze compared to 33.0\% AP at 39ms for YOLOv3 on COCO). Together these contributions for the first time reveal the potential of keypoint-based detection to be useful for applications requiring processing efficiency.  
\end{abstract}

\section{Introduction}
\label{sec:introduction}

Keypoint-based object detection~\cite{tychsen2017denet,wang2017point,law2018cornernet} is a class of methods that generate object bounding boxes by detecting and grouping their keypoints. CornerNet~\cite{law2018cornernet}, the state-of-the-art among them, detects and groups the top-left and bottom-right corners of bounding boxes; it uses a stacked hourglass network~\cite{newell2016stacked} to predict the heatmaps of the corners and then uses associate embeddings~\cite{newell2017associative} to group them. CornerNet allows a simplified design that eliminates the need for anchor boxes~\cite{ren2015faster}, and has achieved state-of-the-art accuracy on COCO~\cite{lin2014microsoft} among single-stage detectors. 

However, a major drawback of CornerNet is its inference speed. It achieves an average precision (AP) of 42.2\% on COCO at an inference cost of 1.1s per image~\footnote{All detectors in this paper are tested on a machine with a 1080Ti GPU and an Intel Core i7-7700k GPU.}, which is too slow for video applications that require real-time or interactive rates. 
Although one can easily speed up inference by reducing the number of pixels processed (e.g.\@ by reducing the number of scales of processing or the image resolution), this can cause a large accuracy drop. For example, single-scale processing combined with reducing the input resolution can speed up the inference of CornerNet to 42ms per image, comparable to the 39ms of the popular fast detector YOLOv3~\cite{redmon2018yolov3}, but would decrease the AP to 25.6\% which is much lower than YOLOv3's 33.0\%. This makes CornerNet less competitive with alternatives in terms of the accuracy-efficiency tradeoff. 

\begin{figure}
    \centering
    \resizebox{0.5\textwidth}{!}{\includegraphics{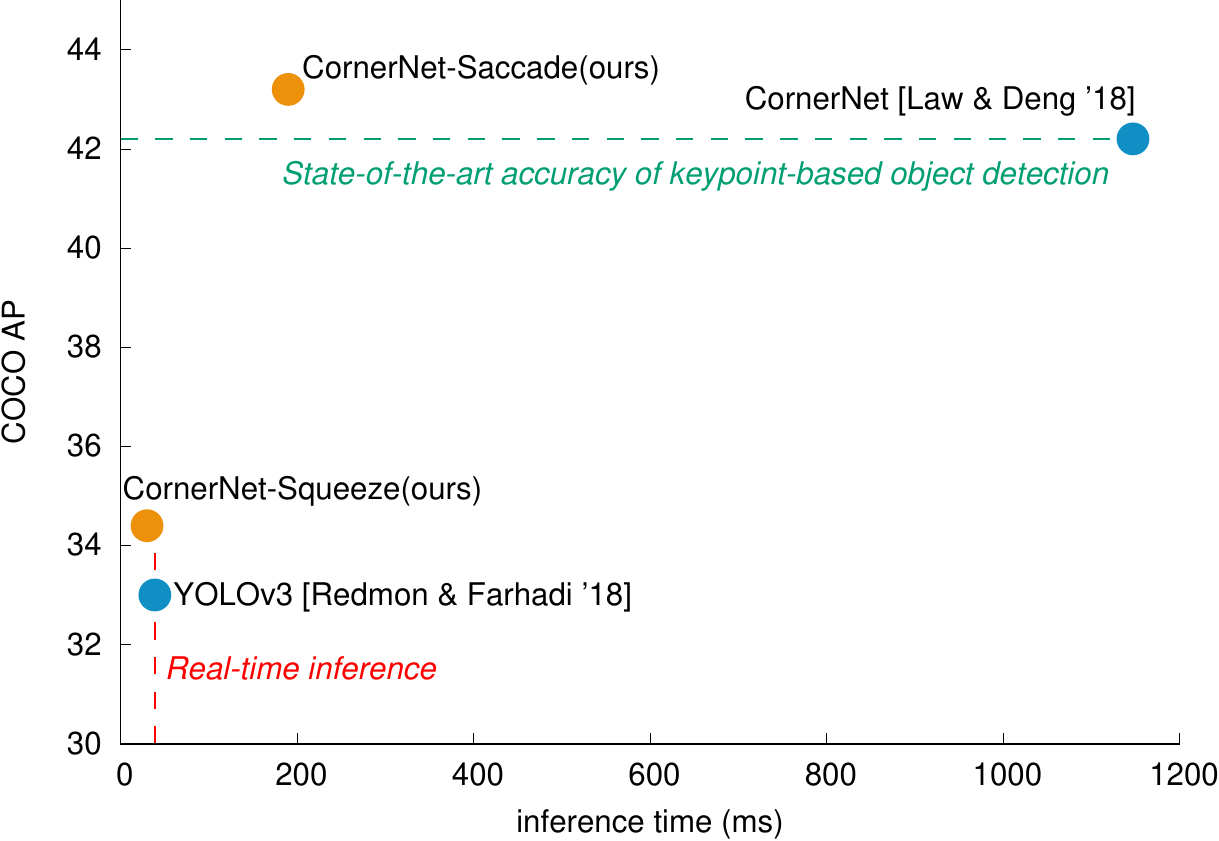}}
    \caption{We introduce CornerNet-Saccade and CornerNet-Squeeze (collectively as CornerNet-Lite), two efficient object detectors based on CornerNet~\cite{law2018cornernet}, a state-of-the-art keypoint based object detector. CornerNet-Saccade speeds up the original CornerNet by 6.0x with a 1\% increase in AP. CornerNet-Squeeze is faster and more accurate than YOLOv3~\cite{redmon2018yolov3}, the state-of-the-art real time detector. All detectors are tested on the same machine with a 1080Ti GPU and an Intel Core i7-7700k CPU.}
    \label{fig:test}
\end{figure}

In this paper we seek to improve the inference efficiency of CornerNet. The efficiency of any object detector can be improved along two orthogonal directions: reducing the number of pixels processed and reducing the amount of processing per pixel. We explore both directions and introduce two efficient variants of CornerNet: \emph{CornerNet-Saccade} and \emph{CornerNet-Squeeze}, which we refer to collectively as \emph{CornerNet-Lite}. 

CornerNet-Saccade speeds up inference by reducing the number of pixels to process. It uses an attention mechanism similar to saccades in human vision~\cite{yarbus2013eye,bahill1975main}. It starts with a downsized full image and generates an attention map, which is then zoomed in on and processed further by the model. This differs from the original CornerNet in that it is applied fully convolutionally across multiple scales. By selecting a subset of crops to examine in high resolution, CornerNet-Saccade improves speed while improving the accuracy. Experiments on COCO show that CornerNet-Saccade achieves an AP of 43.2\% at 190ms per image, a 1\% increase in AP and a 6.0x speed-up over the original CornerNet. 

CornerNet-Squeeze speeds up inference by reducing the amount of processing per pixel. It incorporates ideas from SqueezeNet~\cite{iandola2016squeezenet} and MobileNets~\cite{howard2017mobilenets}, and introduces a new, compact hourglass backbone that makes extensive use of 1$\times$1 convolution, bottleneck layer, and depth-wise separable convolution. With the new hourglass backbone, CornerNet-Squeeze achieves an AP of 34.4\% on COCO at 30ms, simultaneously more accurate and faster than YOLOv3 (33.0\% at 39ms). 

A natural question is whether CornerNet-Squeeze can be combined with saccades to improve its efficiency even further. Somewhat surprisingly, our experiments give a negative answer: \emph{CornerNet-Squeeze-Saccade} turns out slower and less accurate than CornerNet-Squeeze. This is because for saccades to help, the network needs to be able to generate sufficiently accurate attention maps, but the ultra-compact architecture of CornerNet-Squeeze does not have this extra capacity. In addition, the original CornerNet is applied at multiple scales, which provides ample room for saccades to cut down on the number of pixels to process. In contrast, CornerNet-Squeeze is already applied at a single scale due to the ultra-tight inference budget, which provides much less room for saccades to save. 

\smallskip \noindent \textbf{Significance and novelty:} Collectively, these two variants of CornerNet-Lite make the keypoint-based approach competitive, covering two popular use cases: CornerNet-Saccade for offline processing, improving efficiency without sacrificing accuracy, and CornerNet-Squeeze for real-time processing, improving accuracy without sacrificing efficiency.

Both variants of CornerNet-Lite are technically novel. CornerNet-Saccade is the first to integrate saccades with keypoint-based object detection. Its key difference from prior work lies in how each crop (of pixels or feature maps) is processed. Prior work that employs saccade-like mechanisms either detects a single object per crop (e.g.\@ Faster R-CNN~\cite{ren2015faster}) or produces multiple detections per crop with a two-stage network involving additional sub-crops (e.g.\@ AutoFocus~\cite{najibi2018autofocus}). In contrast, CornerNet-Saccade produces multiple detections per crop with a single-stage network. 

CornerNet-Squeeze is the first to integrate SqueezeNet with the stacked hourglass architecture and to apply such a combination on object detection. Prior works that employ the hourglass architecture have excelled at achieving competitive accuracy, but it was unclear whether and how the hourglass architecture can be competitive in terms of efficiency. Our design and results show that this is possible for the first time, particularly in the context of object detection. 

\smallskip \noindent \textbf{Contributions} Our contributions are three-fold: (1) We propose CornerNet-Saccade and CornerNet-Squeeze, two novel approaches to improving the efficiency of keypoint-based object detection; (2) On COCO, we improve the efficiency of state-of-the-art keypoint based detection by 6 fold and the AP from 42.2\% to 43.2\%, (3) On COCO, we improve both the accuracy and efficiency of state-of-the art real-time object detection (to 34.4\% at 30ms from 33.0\% at 39ms of YOLOv3).

\begin{figure}
    \centering
    \resizebox{0.9\textwidth}{!}{\includegraphics[]{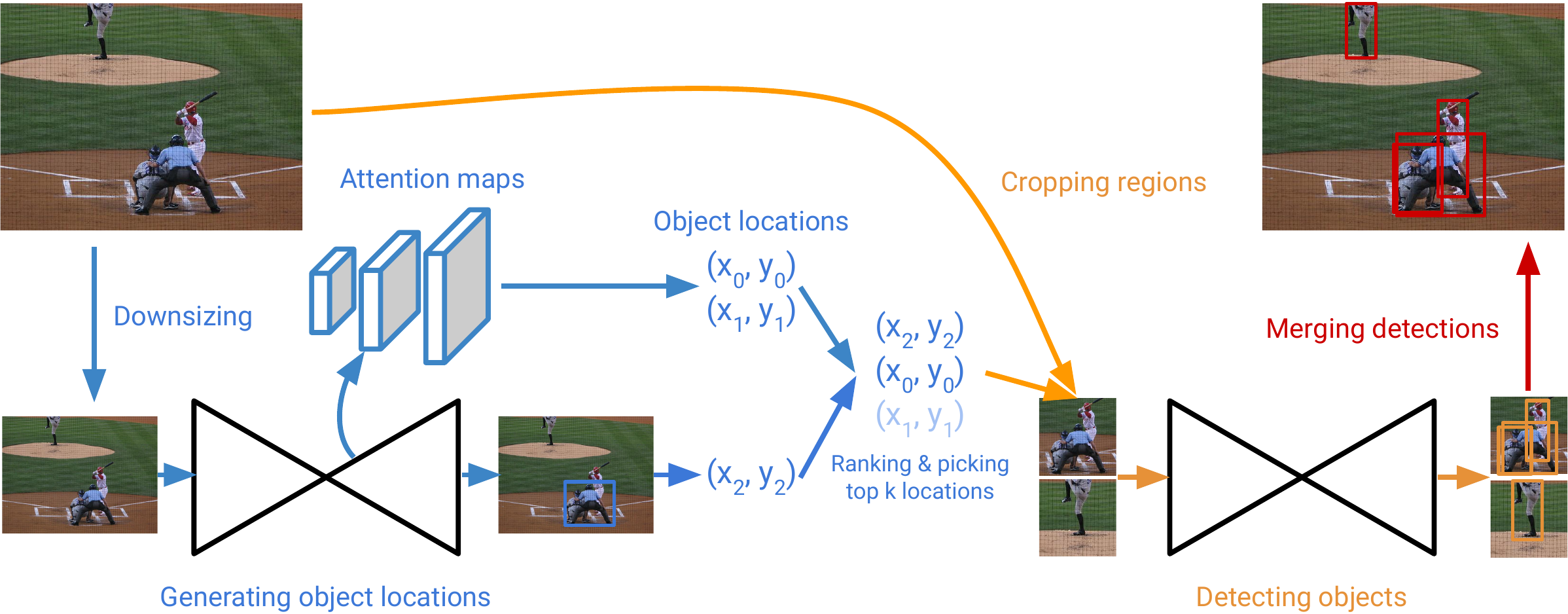}}
    \caption{Overview of CornerNet-Saccade. We predict a set of possible object locations from the attention maps and bounding boxes generated on a downsized full image. We zoom into each location and crop a small region around that location. Then we detect objects in top $k$ regions and merge the detections by NMS.}
    \label{fig:overview}
    \vspace{-3mm}
\end{figure}

\section{Related Work}
\smallskip \noindent \textbf{Saccades in Object Detection.} 
Saccades in human vision refers to a sequence of rapid eye movements to fixate different image regions. In the context of object detection algorithms, we use the term broadly to mean selectively cropping and processing image regions (sequentially or in parallel, pixels or features) during inference. 

There has been a long history of using saccades~\cite{gualdi2011multistage,pedersoli2010recursive,zhang2007real} in object detection to speed up inference. For example, a special case of saccades is a cascade that repeatedly selects a subset of regions for further processing, as exemplified by the Viola-Jones face detector~\cite{viola2001rapid}. 
The idea of saccades has taken diverse forms in various approaches, but can be roughly categorized by how each crop is processed, in particular, what kind of output is produced after processing each crop. 

Saccades in R-CNN~\cite{girshick2014rich}, Fast R-CNN~\cite{girshick2015fast}, and Faster R-CNN~\cite{ren2015faster} take the form of crops representing potential objects. After processing, each crop is either rejected or converted to a single labeled box through classification and regression. Cascade R-CNN~\cite{cai2018cascade} extends Faster R-CNN by using a cascade of classifiers and regressors to iteratively reject or refine each proposal. The saccades in all these R-CNN variants  are thus \emph{single-type and single-object}, in that there is a single type of processing of crops, and each crop produces at most a single object. 

AutoFocus~\cite{najibi2018autofocus}, which builds upon SNIPER~\cite{singh2018sniper} that improved R-CNN training, adds a branch to Faster R-CNN to predict the regions that are likely to contain small objects. Then it applies Faster R-CNN again to each of those regions by cropping the original pixels. In AutoFocus, there are two kinds of cropping, one that can produce multiple objects (by calling Faster R-CNN as a subroutine), and the other that can produce at most a single object (cropping done within Faster R-CNN). The saccades in AutoFocus are thus \emph{multi-type and mixed}, in the sense that two different types of processing are interleaved. 

In contrast, saccades in CornerNet-Saccade are \emph{single-type and multi-object}, in that there is a single type of crop processing and each crop can produce multiple objects at once without additional subcrops. This means that the number of crops processed by CornerNet-Saccade can be much smaller than the number of objects, whereas for R-CNN variants and AutoFocus the number of crops must be no smaller than the number of objects.

\smallskip \noindent \textbf{Efficient Object Detectors.} Other than accuracy~\cite{cai2016unified,shrivastava2016beyond,huang2017speed,lin2017feature,dai2017deformable,zhu2018deformable,he2017mask,peng2018megdet,singh2018analysis,zhai2018feature,xu2018deep,cheng2018revisiting,jiang2018acquisition}, many recent works have improved upon the efficiency of detectors since the introduction of R-CNN~\cite{girshick2014rich}, which applies a ConvNet~\cite{krizhevsky2012imagenet} to 2000 RoIs. Repeatedly applying a ConvNet to the RoIs introduces many redundant computations. SPP~\cite{he2015spatial} and Fast R-CNN~\cite{girshick2015fast} address this by applying a ConvNet fully convolutionally on the image and extracting features directly from the feature maps for each RoI. Faster R-CNN~\cite{ren2015faster} further improves efficiency by replacing the low-level vision algorithm with a region proposal network. R-FCN~\cite{dai2016r} replaces the expensive fully connected sub-detection network with a fully convolutional network, and Light-Head R-CNN~\cite{li2017light} reduces the cost in R-FCN by applying separable convolution to reduce the number of channels in the feature maps before RoI pooling. On the other hand, one-stage detectors~\cite{liu2016ssd,redmon2016you,fu2017dssd,redmon2017yolo9000,lin2017focal,wang2017point,kong2017ron,jeong2017enhancement,zhang2018single,zhao2018m2det,shen2017dsod,zhang2018singleencried} remove the region pooling step of two-stage detectors.

\smallskip \noindent \textbf{Efficient Network Architectures.} The efficiency of ConvNets is important to many mobile and embedded applications. Much attention~\cite{li2016pruning,rastegari2016xnor,zhang2018shufflenet,ma2018shufflenet,laube2018shufflenasnets,hu2018squeeze,sandler2018mobilenetv2} has been given to the design of efficient network architectures. SqueezeNet~\cite{iandola2016squeezenet} proposes a fire module to reduce the number of parameters in AlexNet~\cite{krizhevsky2012imagenet} by 50x, while achieving similar performance.  MobileNets~\cite{howard2017mobilenets} are a class of lightweight networks that use depth-wise separable convolutions~\cite{chollet2017xception}, and proposes strategies to achieve a good trade-off between accuracy and latency. PeleeNet~\cite{wang2018pelee}, in contrast, demonstrates the effectiveness of standard convolutions by introducing an efficient variant of DenseNet~\cite{huang2017densely} consisting of two-way dense layers and a stem block. Other networks were designed specifically for real-time detection. YOLOv2~\cite{redmon2017yolo9000} incorporates ideas from NIN~\cite{lin2013network} to design a new variant of VGG~\cite{simonyan2014very}. YOLOv3~\cite{redmon2018yolov3} further improves DarkNet-19 by making the network deeper and introducing skip connections. RFBNet~\cite{liu2018receptive} proposes a new module which mimics the receptive field of human vision systems to efficiently gather information across different scales.

\section{CornerNet-Saccade}
CornerNet-Saccade detects objects within small regions around possible object locations in an image. It uses the downsized full image to predict attention maps and coarse bounding boxes; both suggest possible object locations. CornerNet-Saccade then detects objects by examining the regions centered at the locations in high resolution. An overview of the pipeline is shown in Fig.~\ref{fig:overview}. 

\subsection{Estimating Object Locations}
The first step in CornerNet-Saccade is to obtain possible object locations in an image. We use downsized full images to predict attention maps, which indicate both the locations and the coarse scales of the objects at the locations. Given an image, we downsize it to two scales by resizing the longer side of the image to $255$ and $192$ pixels. The image of size $192$ is padded with zeros to the size of $255$ so that they can be processed in parallel. There are two reasons for using image at such low resolutions. First, this step should not be a bottleneck in the inference time. Second, the network should easily leverage the context information in the image to predict the attention maps.

For a downsized image, CornerNet-Saccade predicts 3 attention maps, one for small objects, one for medium objects and one for large objects. An object is considered small if the longer side of its bounding box is less than 32 pixels, medium if it is between 32 and 96 pixels, and large if it is greater than 96 pixels\footnote{The sizes are w.r.t the input to the network.}. Predicting locations separately for different object sizes gives us finer control over how much CornerNet-Saccade should zoom in at each location. We can zoom in more at small object locations and less at medium object locations.

We predict the attention maps by using feature maps at different scales. The feature maps are obtained from the backbone network in CornerNet-Saccade, which is an hourglass network~\cite{newell2016stacked}. The feature maps from the upsampling layers in the hourglass are used to predict the attention maps. The feature maps at finer scales are used for smaller objects and the ones at coarser scales are for larger objects. We predict the attention maps by applying a $3 \times 3$ Conv-ReLU module followed by a $1 \times 1$ Conv-Sigmoid module to each feature map. During testing, we only process locations where scores are above a threshold $t$, and we set $t = 0.3$ in our experiments. During training, we set the center of each bounding box on the corresponding attention map to be positive and the rest to negatives. Then we apply the focal loss with $\alpha = 2$.

\subsection{Detecting Objects}
CornerNet-Saccade uses the locations obtained from the downsized image to determine where to process. If we directly crop the regions from the downsized image, some objects may become too small to detect accurately. Hence, we should examine the regions at higher resolution based on the scale information obtained in the first step. 

Based on the locations obtained from the attention maps, we can determine different zoom-in scales for different object sizes: $s_s$ for small objects, $s_m$ for medium objects and $s_l$ for large objects. In general, $s_s > s_m > s_l$ because we should zoom in more for smaller objects, so we set $s_s = 4$, $s_m = 2$ and $s_l = 1$. At each possible location $(x, y)$, we enlarge the downsized image by $s_{i}$, where $i \in \{s, m, l\}$ depending on the coarse object scale. Then we apply CornerNet-Saccade to a $255 \times 255$ window centered at the location.

CornerNet-Saccade uses the same network for attention maps and bounding boxes. Hence, when CornerNet-Saccade processes the downsized image to predict the attention maps, it may already predict boxes for some larger objects in the image. But the boxes may not be accurate. We also zoom into those objects to obatin better boxes.

There are some important implementation details to make processing efficient. First, we process the regions in batch to better utilize the GPU. Second, we keep the original image in GPU memory, and perform resizing and cropping on the GPU to reduce the overhead of transferring image data between CPU and GPU. To further utilize the processing power of the GPU, the cropped images are processed in parallel.

\begin{figure}[t]
    \centering
    \begin{minipage}[t]{0.47\textwidth}
        \centering
        \includegraphics[width=\textwidth]{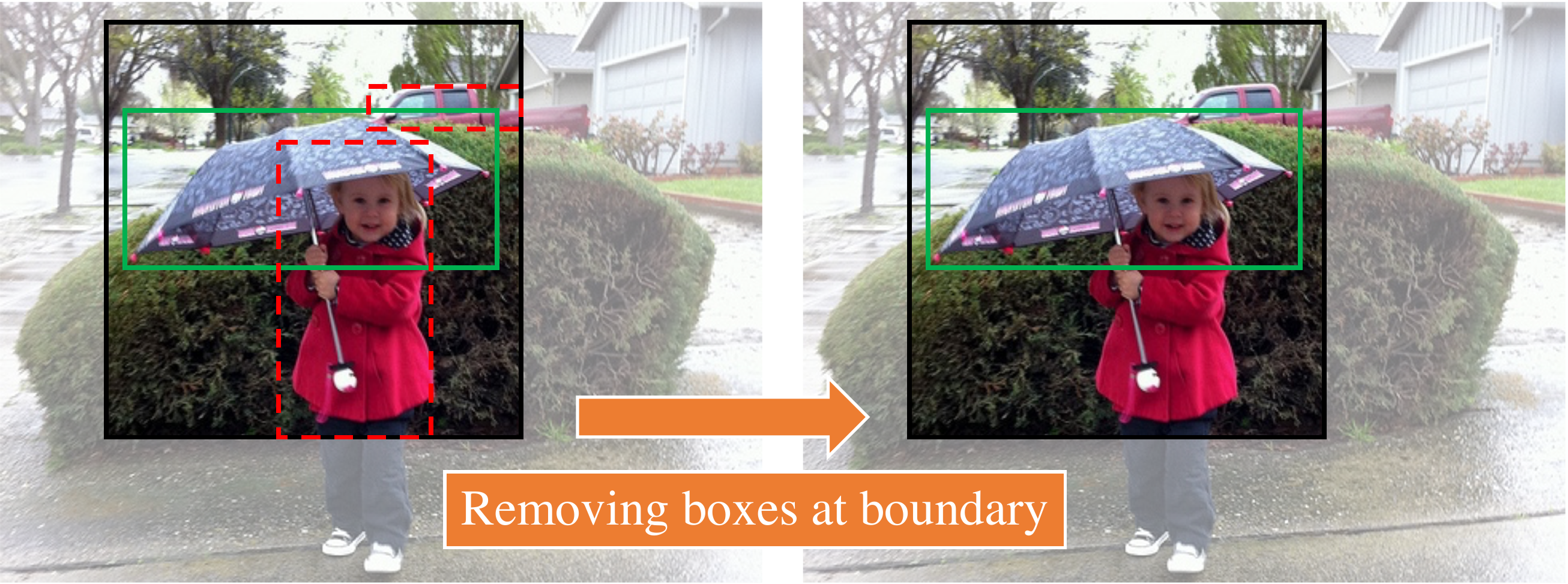}
    \end{minipage}
    \hfill
    \begin{minipage}[t]{0.51\textwidth}
        \centering
        \includegraphics[width=\textwidth]{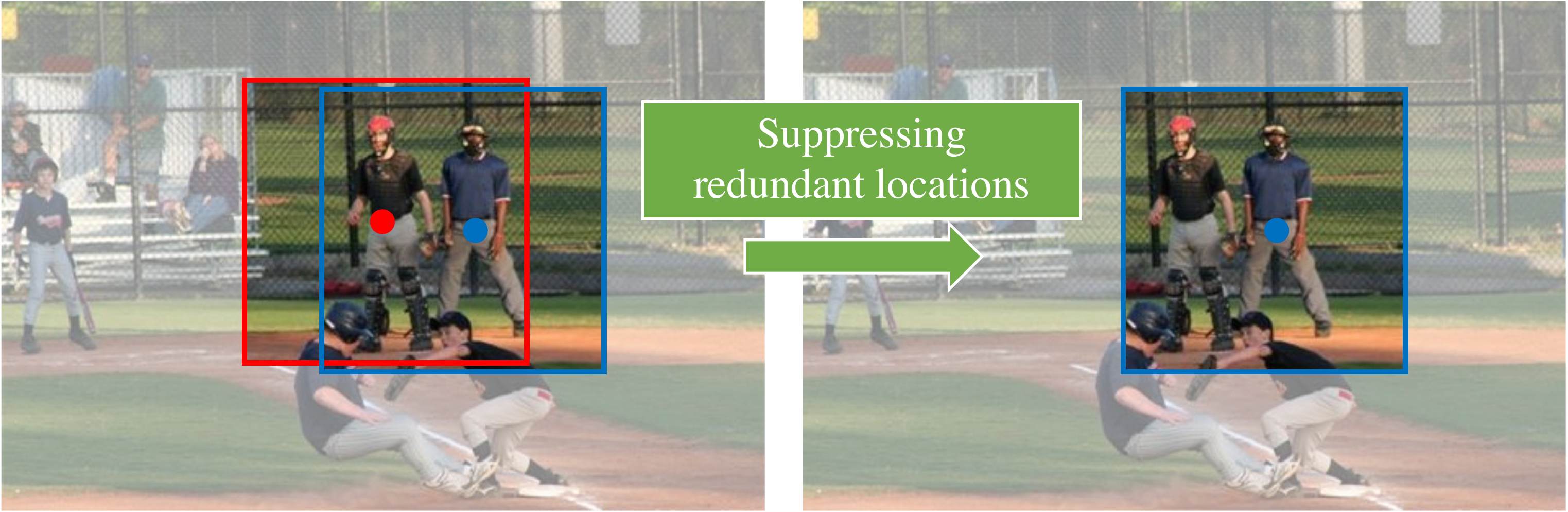}
    \end{minipage}
    \caption{\textbf{Left}: Some objects may not be fully covered by a region. The detector may still generate bounding boxes (red dashed line) for those objects. We remove the bounding boxes which touch the boundaries to avoid such bounding boxes. \textbf{Right}: When the objects are close to each other, we may generate regions that highly overlap with each other. Processing either one of them is likely to detect objects in all highly overlapping regions. We suppress redundant regions to improve efficiency.}
    \label{fig:crop_and_overlap}
    \vspace{-3mm}
\end{figure}

After detecting objects at possible object locations, we merge the bounding boxes and remove redundant ones by applying Soft-NMS~\cite{bodla2017soft}. When we crop the regions, the regions may include parts of the objects at the crop boundaries as shown in Fig.~\ref{fig:crop_and_overlap} (left). The detector may generate bounding boxes for those objects, which may not be removed by Soft-NMS as they may have low overlaps with the bounding boxes of the full objects. Hence, we remove the bounding boxes which touch the crop boundary. During training, we apply the same training losses in CornerNet to train the network to predict corner heatmaps, embeddings and offsets.

When objects are close to each other, we may generate regions that highly overlap with each other as shown in Fig.~\ref{fig:crop_and_overlap} (right). It is undesirable to process both regions as processing either one of them is likely to detect objects in the other. We adopt a procedure similar to NMS to remove redundant locations. First, we rank the object locations, prioritizing locations from bounding boxes over locations from the attention maps. We then keep the best object location and remove the locations that are close to the best location. We repeat the procedure until no object locations are left.

\subsection{Backbone Network}
\label{sec:backbone}
We design a new hourglass backbone network that works better in CornerNet-Saccade. The new hourglass network consists of 3 hourglass modules and has a depth of 54 layers, while Hourglass-104 in CornerNet consists of 2 hourglass modules and has a depth of 104. We refer to the new backbone as Hourglass-54.

Each hourglass module in Hourglass-54 has fewer parameters and is shallower than the one in Hourglass-104. Following the downsizing strategy in Hourglass-104, we downsize the feature by stride 2. We apply one residual module~\cite{he2016deep} after each downsampling layer and in each skip connection. Each hourglass module downsizes the input features 3 times and increases the number of channels along the way (384, 384, 512). There is one residual module with 512 channels in the middle of the module, and one residual module after each upsampling layer. We also downsize the image twice before the hourglass modules.

\subsection{Training Details}
We use Adam~\cite{kingma2014adam} to optimize both the losses for the attention maps and object detection, and use the same training hyperparameters found in CornerNet. The input size to the network is $255 \times 255$, which is also the input resolution during inference. We train the network with a batch size of 48 on four 1080Ti GPUs. In order to avoid over-fitting, we adopt the data augmentation techniques used in CornerNet. When we randomly crop a region around an object, the object is either placed randomly or at the center with some random offset. This ensures that training and testing are consistent as the network detects objects within the crops centered at object locations.

\section{CornerNet-Squeeze}
In contrast to CornerNet-Saccade, CornerNet-Squeeze focuses on reducing the amount of processing per pixel. In CornerNet, most of the computational resources are spent on Hourglass-104. Although Hourglass-104 achieves good performance, it is expensive in terms of number of parameters and inference time. To reduce the complexity of Hourglass-104, we use ideas from SqueezeNet~\cite{iandola2016squeezenet} and MobileNets~\cite{howard2017mobilenets} to design a lightweight hourglass architecture.

\subsection{Ideas from SqueezeNet and MobileNets}
SqueezeNet proposes three strategies to reduce network complexity: (1) replacing $3 \times 3$ kernels with $1 \times 1$ kernels; (2) decreasing input channels to $3 \times 3$ kernels; (3) downsampling late. The building block of SqueezeNet, the fire module, encapsulates the first two ideas. The \textit{fire module} first reduces the number of input channels with a \textit{squeeze} layer consisting of $1 \times 1$ filters. Then, it feeds the result through an \textit{expand} layer consisting of a mixture of $1 \times 1$ and $3 \times 3$ filters.

Based on the insights provided by SqueezeNet, we use the fire module in CornerNet-Squeeze instead of the residual block. Furthermore, inspired by the success of MobileNets, we replace the $3 \times 3$ standard convolution in the second layer with a $3 \times 3$ depth-wise separable convolution, which further improves inference time. Tab.~\ref{tab:squeeze_module} shows a detail comparison between the residual block in CornerNet and the new fire module in CornerNet-Squeeze. 

We do not explore the third idea in SqueezeNet. Since the hourglass network has a symmetrical structure, delayed downsampling results in higher resolution feature maps during the upsampling. Performing convolution on high resolution feature maps is computationally expensive, which would prevent us from achieving real-time detection.

\begin{table}[ht]
    \centering
    \def\arraystretch{1.1}
    \resizebox{0.6\textwidth}{!}{
    \begin{tabular}{c|c|c}
    Input & Operator & Output \\ \Xhline{3\arrayrulewidth} 
    \multicolumn{3}{l}{Residual block in CornerNet} \\ \hline
    $h \times w \times k$ & $3 \times 3$ Conv, ReLU & $h \times w \times k'$ \\
    $h \times w \times k'$ & $3 \times 3$ Conv, ReLU & $h \times w \times k'$ \\ \Xhline{3\arrayrulewidth}
    \multicolumn{3}{l}{Fire module in CornerNet-Squeeze} \\ \hline
    $h \times w \times k$ & $1 \times 1$ Conv & $h \times w \times \frac{k'}{2}$ \\
    $h \times w \times \frac{k'}{2}$ & $1 \times 1$ Conv + $3 \times 3$ Dwise, ReLU & $h \times w \times k'$ \\ \Xhline{3\arrayrulewidth}
    \end{tabular}}
    \caption{Comparison between the residual block and the new fire module.}
    \label{tab:squeeze_module}
\end{table}

Other than replacing the residual blocks, we also make a few more modifications. We reduce the maximum feature map resolution of the hourglass modules by adding one more downsampling layer before the hourglass modules, and remove one downsampling layer in each hourglass module. CornerNet-Squeeze correspondingly downsizes the image three times before the hourglass module, whereas CornerNet downsizes the image twice. We replace the $3 \times 3$ filters with $1 \times 1$ filters in the prediction modules of CornerNet. Finally, we replace the nearest neighbor upsampling in the hourglass network with transpose convolution with a $4 \times 4$ kernel.

\subsection{Training Details}
\label{sec:cornernet_squeeze_details}
We use the same training losses and hyperparameters of CornerNet to train CornerNet-Squeeze. The only change is the batch size. Downsizing the image one more time prior to the hourglass modules reduces the memory usage by 4 times under the same image resolution in CornerNet-Squeeze. We train the network with a batch size of $55$ on four 1080Ti GPUs (13 images on the master GPU and 14 images per GPU for the rest of the GPUs).

\section{Experiments}
\smallskip \noindent \textbf{Implementation Details.} CornerNet-Lite is implemented in PyTorch~\cite{paszke2017automatic}. We use COCO~\cite{lin2014microsoft} to evaluate CornerNet-Lite and compare it with other detectors. To measure the inference time, for each detector, we start the timer as soon as it finishes reading the image and stop the timer as soon as it obtains the final bounding boxes. To provide fair comparisons between different detectors, we measure the inference times on the same machine with a 1080Ti GPU and an Intel Core i7-7700k CPU.

\begin{table}[h]
    \centering
    \resizebox{0.7\textwidth}{!}{
    \begin{tabular}{l|c|c|c}
    Detector          & GPU              & Quantity & Total Mem \\ \Xhline{3\arrayrulewidth}
    CornerNet         & Titan X (PASCAL) & 10       & 120GB     \\
    CornerNet-Saccade & 1080Ti           & 4        & 44GB      \\ \Xhline{3\arrayrulewidth}
    \end{tabular}}
    \caption{CornerNet-Saccade saves more than 60\% GPU memory and requires only 4 GPUs to train, while it outperforms CornerNet.}
    \label{tab:training_efficiency}
    \vspace{-3mm}
\end{table}

\smallskip \noindent \textbf{Training Efficiency of CornerNet-Saccade.} CornerNet-Saccade not only improves the efficiency in testing but also in training. Tab.~\ref{tab:training_efficiency} shows that we are able to train CornerNet-Saccade on only four 1080Ti GPUs with a total of 44GB GPU memory while CornerNet requires ten Titan X (PASCAL) GPUs with a total of 120GB GPU memory. We reduce the memory usage by more than 60\%. Neither CornerNet nor CornerNet-Saccade uses mixed precision training~\cite{micikevicius2017mixed}.

\smallskip \noindent \textbf{Performance Analysis of Hourglass-54 in CornerNet-Saccade.} We introduce a new hourglass, Hourglass-54, in CornerNet-Saccade, and perform two experiments to better understand the performance contribution of Hourglass-54. First, we train CornerNet-Saccade with Hourglass-104 instead of Hourglass-54. Second, we train CornerNet with Hourglass-54 instead of Hourglass-104. For the second experiment, due to limited resources, we train both networks with a batch size of 15 on four 1080Ti GPUs and follow the training details in~\cite{law2018cornernet}. 

\begin{table}[h]
    \centering
    \resizebox{0.65\textwidth}{!}{
    \begin{tabular}{l|c|c|c|c|c}
                                & AP   & $\text{AP}^{s}$ & $\text{AP}^{m}$ & $\text{AP}^{l}$ & $\text{AP}^{att}$ \\ \Xhline{3\arrayrulewidth}
    CornerNet-Saccade w/ HG-54  & 42.6 & 25.5            & 44.3            & 58.4            & 42.7 \\
    + gt attention              & 50.3 & 32.3            & 53.4            & 65.3            & -    \\
    CornerNet-Saccade w/ HG-104 & 41.4 & 23.8            & 43.5            & 57.1            & 40.1 \\
    + gt attention              & 48.9 & 32.4            & 51.8            & 62.6            & -    \\
    \Xhline{3\arrayrulewidth}
    CornerNet w/ HG-54          & 37.2 & 18.4            & 40.3            & 49.0            & - \\
    CornerNet w/ HG-104         & 38.2 & 18.6            & 40.9            & 50.1            & - \\ \Xhline{3\arrayrulewidth}
    \end{tabular}}
    \caption{Hourglass-54 produces better results when combined with saccade on the COCO validation set.}
    \label{tab:hourglass_54}
\end{table}

Tab.~\ref{tab:hourglass_54} shows that CornerNet-Saccade with Hourglass-54 (42.6\% AP) is more accurate than with Hourglass-104 (41.4\%). To investigate the difference in performance, we evaluate the quality of both the attention maps and bounding boxes. First, predicting the attention maps can be seen as a binary classification problem, where the object locations are positives and the rest are negatives. We measure the quality of the attention maps by average precision, denoted as $\text{AP}^{att}$. Hourglass-54 achieves an $\text{AP}^{att}$ of 42.7\%, while Hourglass-104 achieves 40.1\%, suggesting that Hourglass-54 is better at predicting attention maps. 

Second, to study the quality of bounding boxes from each network, we replace the predicted attention maps with the ground-truth attention maps, and also train CornerNet with Hourglass-54. With the ground-truth attention maps, CornerNet-Saccade with Hourglass-54 achieves an AP of 50.3\% while CornerNet-Saccade with Hourglass-104 achieves an AP of 48.9\%. CornerNet with Hourglass-54 achieves an AP of 37.2\%, while Hourglass-104 achieves 38.2\%. The results suggest that Hourglass-54 produces better bounding boxes when combined with saccade.

\begin{table}
    \centering
    \resizebox{0.65\textwidth}{!}{
    \begin{tabular}{l|c|c}
                          & Time & AP   \\ \Xhline{3\arrayrulewidth}
         CornerNet         & 211ms & 31.4* \\
         + w/o flipped image & 111ms & 29.7* \\
         + one extra downsampling before HG modules & 41ms & 33.0 \\ 
         + replace residual blocks with new fire modules & 31ms & 29.8 \\ 
         + replace $3 \times 3$ with $1 \times 1$ conv in prediction layers & 28ms & 29.8 \\
         + upsample using transpose conv (CornerNet-Squeeze) & 30ms & 30.3 \\
         \Xhline{3\arrayrulewidth}
         replace new fire modules with original fire modules~\cite{iandola2016squeezenet} & 41ms & 31.0 \\
         replace new fire modules with MobileNets~\cite{howard2017mobilenets} modules & 35ms & 26.8 \\
         \Xhline{3\arrayrulewidth}
         replace squeezed hourglass with SqueezeNet~\cite{iandola2016squeezenet} & 24ms & 12.5 \\
         replace squeezed hourglass with MobileNets~\cite{howard2017mobilenets} & 26ms & 22.3 \\
         replace squeezed hourglass with ShuffleNet~\cite{zhang2018shufflenet} & 31ms & 19.7 \\
         \Xhline{3\arrayrulewidth}
    \end{tabular}}
    \caption{Ablation study on CornerNet-Squeeze. *Note that CornerNet is trained with a smaller batch size.}
    \label{tab:cornernet_squeeze_ablation}
\end{table}

\smallskip \noindent \textbf{CornerNet-Squeeze Ablation Study.} We study each major change in CornerNet-Squeeze to understand its contribution to the inference time and AP, and compare squeezed hourglass network to different networks in Tab.~\ref{tab:cornernet_squeeze_ablation}. To conserve GPUs, each model is only trained for 250k iterations, following the details in Sec.~\ref{sec:cornernet_squeeze_details}. Results in Tab.~\ref{tab:cornernet_squeeze_ablation} show that our squeezed hourglass network achieves better performance and efficiency than existing networks.

\begin{table}
    \centering
    \resizebox{0.65\columnwidth}{!}{
    \begin{tabular}{l|c|c|c|c|c}
                                & Time & AP   & $\text{AP}^{s}$ & $\text{AP}^{m}$ & $\text{AP}^{l}$\\ \Xhline{3\arrayrulewidth}
    CornerNet-Squeeze-Saccade   & 61ms & 32.7 & 17.3            & 32.6            & 47.1 \\
    + gt attention              & -    & 38.0 & 24.4            & 39.3            & 50.2 \\
    \hline
    CornerNet-Squeeze           & 30ms & 34.4 & 14.8            & 36.9            & 49.5 \\ \Xhline{3\arrayrulewidth}
    \end{tabular}}
    \caption{Saccade only helps if the attention maps are sufficiently accurate.}
    \label{tab:cornernet_squeeze_saccade}
    \vspace{-3mm}
\end{table}

\smallskip \noindent \textbf{CornerNet-Squeeze-Saccade.} We try combining CornerNet-Squeeze with saccades to further improve the efficiency. The results in Tab.~\ref{tab:cornernet_squeeze_saccade} suggest that saccade can only help if the attention maps are sufficiently accurate. Due to its ultra-compact architecture, CornerNet-Squeeze-Saccade does not have enough capacity to detect objects and predict attention maps simultaneously. Furthermore, CornerNet-Squeeze only operates on single scale images, which provides much less room for CornerNet-Squeeze-Saccade to save. CornerNet-Squeeze-Saccade may process more number of pixels than CornerNet-Squeeze, slowing down the inference time.

\begin{table}[h]
    \centering
    \resizebox{0.65\columnwidth}{!}{
    \begin{tabular}{l|c|c|c|c|c}
                        & Time   & AP   & $\text{AP}^{s}$ & $\text{AP}^{m}$ & $\text{AP}^{l}$\\ \Xhline{3\arrayrulewidth}
    YOLOv3              & 39ms   & 33.0 & 18.3            & 35.4            & 41.9 \\
    CornerNet-Squeeze   & 30ms   & 34.4 & 13.7            & 36.5            & 47.4 \\
    \Xhline{3\arrayrulewidth}
    CornerNet (single)  & 211ms  & 40.6 & 19.1            & 42.8            & 54.3 \\
    CornerNet (multi)   & 1147ms & 42.2 & 20.7            & 44.8            & 56.6 \\
    CornerNet-Saccade   & 190ms  & 43.2 & 24.4            & 44.6            & 57.3 \\
    \Xhline{3\arrayrulewidth}
    \end{tabular}}
    \caption{CornerNet-Lite versus CornerNet and YOLOv3 on COCO test set.}
    \label{tab:coco_test}
\end{table}

\smallskip \noindent \textbf{Comparisons with YOLO and CornerNet.} We also compare CornerNet-Lite with CornerNet and YOLOv3 on COCO test set in Tab.~\ref{tab:coco_test}. CornerNet-Squeeze is faster and more accurate than YOLOv3. CornerNet-Saccade is more accurate than CornerNet at multi-scales and 6 times faster.

\section{Conclusions}
We propose CornerNet-Lite which is a combination of two efficient variant of CornerNet: CornerNet-Saccade and CornerNet-Squeeze. Together these contributions for the first time reveal the potential of keypoint-based detection to be useful for applications requiring processing efficiency.

\smallskip \noindent \textbf{Acknowledgements} This work is partially supported by a DARPA grant FA8750-18-2-0019 and an IARPA grant D17PC00343.

\bibliography{egbib}
\end{document}